\documentclass[10pt,twocolumn,letterpaper]{article}

\usepackage{wacv}              

\usepackage[accsupp]{axessibility}
\usepackage{graphicx}
\usepackage{amsmath}
\usepackage{amssymb}
\usepackage{booktabs}
\usepackage{scalerel}
\usepackage{amsmath}
\usepackage{multirow}
\usepackage{algorithm}
\usepackage{algpseudocode}
\usepackage{breqn}
\usepackage[mathscr]{euscript}
\usepackage{pifont}

\usepackage{color, colortbl}

\usepackage{amsmath,amsfonts,bm}

\def\eqref#1{equation~\ref{#1}}

\def\1{\bm{1}}

\DeclareMathAlphabet{\mathsfit}{\encodingdefault}{\sfdefault}{m}{sl}
\SetMathAlphabet{\mathsfit}{bold}{\encodingdefault}{\sfdefault}{bx}{n}

\newcommand{\cmark}{\ding{51}}%
\newcommand{\xmark}{\ding{55}}%

\definecolor{Gray}{gray}{0.9} 
\definecolor{maroon}{rgb}{0.5, 0.0, 0.0}

\definecolor{arylideyellow}{rgb}{0.91, 0.84, 0.42}
\definecolor{burntorange}{rgb}{0.8, 0.33, 0.0}
\definecolor{amber}{rgb}{1.0, 0.75, 0.0}
\definecolor{bananamania}{rgb}{0.98, 0.91, 0.71}
\definecolor{blond}{rgb}{0.98, 0.94, 0.75}
\definecolor{LightCyan}{rgb}{0.88,1,1}

\newlength\myindent
\graphicspath{{./}{./figures/}}
\def\Vec#1{{\boldsymbol{#1}}}

\usepackage[pagebackref,breaklinks,colorlinks]{hyperref}

\usepackage[capitalize]{cleveref}
\crefname{section}{Sec.}{Secs.}
\Crefname{section}{Section}{Sections}
\Crefname{table}{Table}{Tables}
\crefname{table}{Tab.}{Tabs.}

\def\wacvPaperID{1461} 
\def\confName{WACV}
\def\confYear{2024}

\begin{document}

\title{On Manipulating Scene Text in the Wild with Diffusion Models}

\author{Joshua Santoso\\
Independent Researcher\\
{\tt\small janojoshua@gmail.com}
\and
Christian Simon\\
The Australian National University\\
{\tt\small christian.simon@anu.edu.au}
\and
Williem\\
Independent Researcher\\
{\tt\small williem.pao@gmail.com}
}
\maketitle

\begin{abstract}
   Diffusion models have gained attention for image editing yielding impressive results in text-to-image tasks. On the downside, one might notice that generated images of stable diffusion models suffer from deteriorated details. This pitfall impacts image editing tasks that require information preservation \eg, scene text editing. As a desired result, the model must show the capability to replace the text on the source image to the target text while preserving the details \eg, color, font size, and background. To leverage the potential of diffusion models, in this work, we introduce  \textbf{D}iffusion-\textbf{B}as\textbf{E}d \textbf{S}cene \textbf{T}ext manipulation Network so-called DBEST. Specifically, we design two adaptation strategies, namely one-shot style adaptation and text-recognition guidance. In experiments, we thoroughly assess and compare our proposed method against state-of-the-arts on various scene text datasets, then provide extensive ablation studies for each granularity to analyze our performance gain. Also, we demonstrate the effectiveness of our proposed method to synthesize scene text indicated by competitive Optical Character Recognition (OCR) accuracy. Our method achieves 94.15\% and 98.12\% on COCO-text and ICDAR2013 datasets for character-level evaluation.
\end{abstract}


\section{Introduction}
\label{sec:intro}
    Scene text manipulation has gained significant attention in computer vision due to its practical applications. The promising application of scene text manipulation is real-time sign translation allowing for instant translation of signs in various languages~\cite{FragosoWACV2011}. Additionally, scene text manipulation can be used to protect privacy by obscuring sensitive information on images~\cite{ZhangAAAI2019}.
    
In deep learning, synthetic scene texts can be used to augment data for various training purposes \eg, text classification and detection. 
For instance, Tang~\etal~\cite{StrokeErase2021tang} used a synthesized dataset yielding an improved model capability for detecting and erasing text in the wild. Though the synthetic data can be collected from image editors using editing software, this labor task is costly and requires expertise to generate consistent results. Thus, automatic tools for scene text are highly desirable. However, developing such tools is not easy, especially to ensure the edited text compatible for further processing \eg, preserving the details to avoid performance degradation on Optical Character Recognition (OCR) models. 

\begin{figure}[t]
    \centering
        \includegraphics[width=1.\linewidth]{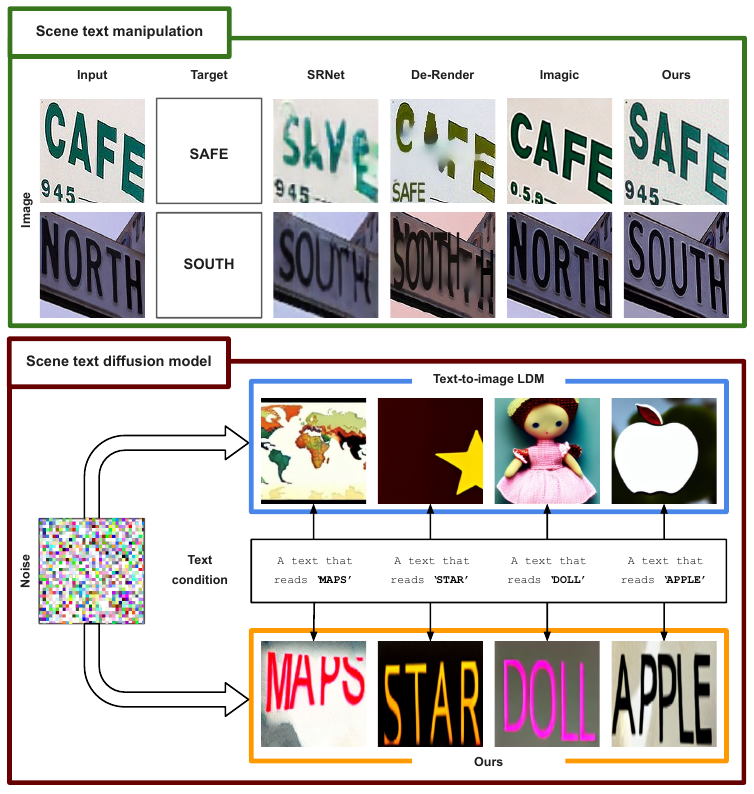}
   \caption{Top: Comparison between state-of-the-art methods and our method from given input image and target text on scene text manipulation. 
   Bottom: Comparison between baseline text-to-image Latent Diffusion Model (LDM)~\cite{RombachCVPR2022} represented with \textit{{\color{blue} blue box}} and our method represented with~\textit{{\color{maroon} red box}} on scene text domain from given random noise and text condition as an input. 
   }
   \label{fig:problem}
\end{figure}

    The automatic way for scene text editing can be achieved using deep generative models~\cite{BianCVM2022, ShimodaICCV2021, WuACMM2019, YangCVPR2020, YangICCV2019, ZhangAAAI2019}. 
    As Generative Adversarial Networks (GANs)~\cite{GoodfellowNIPS2014} progressing, earlier studies have addressed scene text manipulation~\cite{ShimodaICCV2021, WuACMM2019, YangCVPR2020} exploiting the generator and discriminator scheme. 
    Wu \etal~\cite{WuACMM2019} demonstrated the ability of GANs by formulating the problem as an image style transfer task, where the models employ the target text while imposing the styles from the original text. Shimoda \etal~\cite{ShimodaICCV2021} propose to train a text vectorization model and parse text components (\eg style, background, and color) for editing. While effective, the performance of these prior methods to edit the scene text is still unsatisfactory in challenging cases, as shown in Fig.~\ref{fig:problem}. In inference, these prior techniques depend heavily on additional modules \eg character recognition models and synthetic text generators. Moreover, the underlying generative models of these prior works (\ie GANs) have limited capacity to manipulate text with a high degree of variability and complexity.  
    
In this paper, we exploit the diffusion models~\cite{HoNIPS2020, RombachCVPR2022, SongICLR2021}, which have shown impressive results for text-to-image tasks, an alternative competing method for synthesizing scene texts. Also, compared to GAN solutions, diffusion models do not suffer from training instability and mode-collapse~\cite{HoNIPS2020}. Especially, the recent advances in diffusion models allow for incorporating multi-modal conditional inputs \eg, generating an image using a text condition known as text-to-image models~\cite{NicholICML2022, RombachCVPR2022, RameshICML2021}.
These works, especially Latent Diffusion Model (LDM)~\cite{RombachCVPR2022}, provide off-the-shelf pretrained-models for generating images from text prompts. Several works (\eg, Imagic~\cite{KawarArXiv2022} and Null-Inversion~\cite{RonArXiv2022}) have explored the possibility of using the pretrained models for image editing while preserving parts of interests. Even though these prior methods have shown remarkable results for general image editing, implementing this to scene text editing is not straightforward. For example, Fig.~\ref{fig:problem} (top) shows that the text scene cannot be manipulated using Imagic~\cite{KawarArXiv2022}. Instead of generating scene texts, the prior methods tend to generate scenes with some objects related to the text prompt, as illustrated in Fig.~\ref{fig:problem} (bottom). In contrast to the aforementioned image editing models, our work focuses on editing scene texts while preserving fine details of text information. 
In this work, to manipulate text in the wild, we propose a \textbf{D}iffusion-\textbf{B}as\textbf{E}d \textbf{S}cene \textbf{T}ext manipulation Network so-called DBEST. Our design uses text prompts; thus, we have the benefit of leveraging the large language model as a condition for image editing. In summary, we list our contributions below:

\begin{itemize}
    \item{ We identify the limitation of existing diffusion-based image editing models and introduce a diffusion model akin to the text-inversion technique that performs scene text manipulation. For image generation, our model has no modification over the original LDM, thus allowing easy integration into other text-related pipelines. }
    \item{ We present a synthetic dataset as a part of our training strategy, omitting the cost to collect the scene text data.}
    
    \item {We introduce the concept of one-shot style adaptation for scene text manipulation, which aims to maintain the source style on the edited image. }
    
    \item {We devise the classifier guidance approach using a text recognition model, yielding a significant improvement in terms of quality and quantity. To highlight, the OCR word accuracy of our proposed network achieves 84.83\% on the SynText dataset leading to 13\% improvements over the state-of-the-art performance.   
    }
\end{itemize}

\section{Related works}

\paragraph{Scene text manipulation.} 
The manipulation of scene text can be divided into three sub-fields: text effect transfer, text removal, and text manipulation. Text effect transfer involves transferring a style effect from a source style effect image to a target scene text. Text removal involves removing text from a scene while maintaining the scene's appearance. Finally, text manipulation entails editing a source text image with desired text while preserving the text style and background from the source image.

Specific to text manipulation, STEFFAN~\cite{RoyCVPR2020} introduces a character-level editing text where the input image is first converted into a per-character mask. SRNet~\cite{WuACMM2019} introduces a GAN-Based approach for word-level editing where the desired text is first converted to image level. Then, the target text image can be edited through three sequence steps: word geometry transformation, background retrieval, and fusion network. SwapText~\cite{YangCVPR2020} is the extension of SRNet, where the word geometry transformation is replaced by per-character geometry transformation. De-Render~\cite{ShimodaICCV2021} introduces a new approach where the input image is first fed to the rendering parameters prediction model. Then, the reconstruction model uses the predicted parameters to reconstruct the text scene image. Finally, the text can be edited by updating rendered parameter. 

Our method shares similarities with SRNet, but while SRNet struggles with editing images using the style transfer concept, we aim to explore techniques such as conditional GAN~\cite{OdenaICML2017}. Unlike converting the target text to the image level, our method operates at the text level for more efficient editing.  


\paragraph{Diffusion model for image manipulation.}
Diffusion model has shown its capability to beat GAN on image generation. From the given image, several works show impressive work for image-to-image translation~\cite{HoNIPS2020, RombachCVPR2022, SongICLR2021}. In further research, It is also possible to control the generated image from the given condition as proven by~\cite{AvrahamiCVPR2022, KawarArXiv2022, KimCVPR2022, MengICLR2022, RonArXiv2022, RombachCVPR2022}. In brief, using the pre-trained language-vision model such as CLIP~\cite{PatashnikICCV2021, CrowsonCLIPHQ}, Kim~\etal~\cite{KimCVPR2022} shows the excellent result in manipulating a whole image by using CLIP as a loss function. Avrahami~\etal~\cite{AvrahamiCVPR2022} introduces local editing guided CLIP loss where the local mask is necessary as the additional input. 

Alternatively, Kawar~\etal~\cite{KawarArXiv2022} and Ron~\etal~\cite{RonArXiv2022} utilize the pre-trained model from Latent diffusion model (LDM)~\cite{RombachCVPR2022} and introduce the fine-tune approach. First, Ron~\etal fine-tuned the unconditional embedding produced by LDM to maintain the style of the input image. Then, the editing process can be done using Prompt-to-prompt (p2p)~\cite{HertzArXiv2022}. Kawar~\etal proposed two steps: conditional embedding optimization and diffusion model fine-tuning with the optimized embedding. Then, the edited image is generated by interpolating the original and optimized embedding during sampling. 

However, we investigate that the success of the methods above relies on language-vision models such as CLIP or LDM. While those methods perform well on the text-to-image, they still fail on the text to scene text domain, as shown in Fig.~\ref{fig:problem}. Therefore, our approach utilizes LDM pre-trained model on text-to-image as our base model. Then, we train the model with a text-oriented image dataset to shift the domain to text-to-scene-text as shown in Fig.~\ref{fig:problem}.  

\vspace{-0.1cm}
\section{Preliminaries}
  Inspired by the success of the diffusion models~\cite{HoNIPS2020, RombachCVPR2022, SongICLR2021} in generating high-quality images from a given prompt (text to image), our method employs the concept of diffusion process to generate a sample by multiple denoising steps.  
  Specifically, our model is built on memory-efficient diffusion models \ie the Latent Diffusion Model (LDM), and focuses on performing diffusion for text to scene text generation in the latent space. In this section, we briefly overview the fundamental principles of LDM.
\vspace{-0.3cm}
\paragraph{Diffusion models.}
Diffusion probabilistic models~\cite{HoNIPS2020, DicksteinICML2015} mainly consist of two processes: 1) forward steps and 2) reverse steps. Starting from the distribution of an input image $q(\Vec{x}_0)$, where $\Vec{x}_0 \in \mathbb{R}^{3\times H \times W}$ in the RGB space, the forward steps $q(\Vec{x}_{1:T}|\Vec{x}_0)$ are a sequence of the Markov chain that outputs a series of noisy image $\Vec{x}_1, \cdots , \Vec{x}_T$.  The Gaussian noise $\epsilon$ is gradually added to the input image $\Vec{x}_0$ based on a variance schedule $\{\beta_1, \cdots , \beta_T\}$.  The forward process is defined as:
\begin{align}
\begin{split}
        q(\Vec{x}_t|\Vec{x}_{t-1}) &= \mathcal{N}(\Vec{x}_t; \sqrt{1 - \beta_t}\Vec{x}_{t-1}, \beta_t, \pmb{I}).
\end{split}
\end{align}
As the noisy image ${\Vec{x}_T}$ must be reconstructed to ${\Vec{x}_0}$, we need to learn a model using the reverse process estimating the joint distribution $p_{\theta}(\Vec{x}_{0:T})$ yielding: 
\begin{align}
\begin{split}
        p_{\theta}(\Vec{x}_{0:T}) &= p(\Vec{x}_T)\prod_{t=1}^T p_{\theta}(\Vec{x}_{t-1}|x_t), \\
        p_{\theta}(\Vec{x}_{t-1}|\Vec{x}_t) &= \mathcal{N}(\Vec{x}_{t-1}; \mu_{\theta}(\Vec{x}_t, t), \Sigma_{\theta}(\Vec{x}_t, t)).
    \label{eq:rev_ddpm}
 \end{split}   
\end{align}
Alternatively, the noised image $x_t$ can be sampled directly from $x_0$. With $\alpha_t := 1 - \beta_t $ and $\alpha_{\bar{t}} := \prod_{s=0}^T \alpha_s$, $q(\Vec{x}_t|\Vec{x}_0)$ can be formulated as: 
\begin{equation}
    q(\Vec{x}_t|\Vec{x}_0) = \mathcal{N}(\Vec{x}_t; \sqrt{\alpha_t^{\_}}\Vec{x}_0, (1 - \alpha_{\bar{t}}) \mathbf{I}).
\label{eq:fwd_ddpm}
\end{equation}
Ho~\etal~\cite{HoNIPS2020} introduces the approximation noise model $\epsilon_{\theta}(\Vec{x}_t, t)$ to predict $\mu_{\theta}(\Vec{x}_t, t)$: 
\begin{equation}
    \mu_{\theta}(\Vec{x}_t, t) = \dfrac{1}{\sqrt{\alpha_t}} \left( \Vec{x}_t - \dfrac{1 - \alpha_t}{\sqrt{1 - \alpha_{\bar{t}}}} \epsilon_{\theta}(\Vec{x}_t, t) \right).
\end{equation}
\paragraph{Diffusion in the latent space.} LDM  has two modules: perceptual compression and denoising U-Net. The perceptual compression is able to encode the given $\Vec{x}$ into a latent space $\Vec{z} = \scaleto{\varepsilon}{7pt}(\Vec{x})$ and is able to reconstruct back from latent to image space $\Vec{x}= \scaleto{\mathcal{D}}{7pt}(\scaleto{\varepsilon}{7pt}(\Vec{x}))$. The diffusion model $e_{\theta}(\Vec{z}_t, t)$ generates the predicted noise of the noisy latent image $\Vec{z}_t$ at time $t$. The loss function can be formulated as 
\begin{equation}
   \mathcal{L}_{\mathrm{LDM}}:=\mathbb{E}_{\Vec{z}, \epsilon\sim\mathcal{N}(0,1),t}  \big\|\epsilon - \epsilon_{\theta}( \Vec{z}_t, t )\big\|_2^2   .
\label{eq:ldm}
\end{equation}
If the condition (\eg text, image, or label)  $y$ exists, then the loss function is presented as
\begin{equation}
\mathcal{L}_{\mathrm{LDM}}:=\mathbb{E}_{\Vec{z}, \Vec{y}, \epsilon\sim\mathcal{N}(0,1),t} \big\|\epsilon - \epsilon_{\theta}(\Vec{z}_t, t, \tau_{\psi}(\Vec{y}) )\big\|_2^2 ,
\label{eq:cldm}
\end{equation}
 where $\tau_{\psi}$ is a conditional model, \eg, BERT~\cite{DevlinNAACL2019} or CLIP~\cite{RadfordICML2021}, that takes $\Vec{y}$ as an input and generates an embedding feature $\Vec{e} \in \mathbb{R}^{k\times d}$. Here, $k$ denotes the number of tokens in $\Vec{y}$, and $d$ represents the feature dimension. This embedding feature then is combined with the latent vector $\Vec{z}$  using a cross attention mechanism in the denoising model $\epsilon_{\theta}$. Note that we use the identical conditioning mechanisms in LDM~\cite{RombachCVPR2022} with the cross-attention mechanism of multi-modalities in the intermediate layers of the U-Net~\cite{RonnebergerMICCAI2015}.

\begin{figure*}[t]  
    \begin{center}
        \includegraphics[width=1.\linewidth]{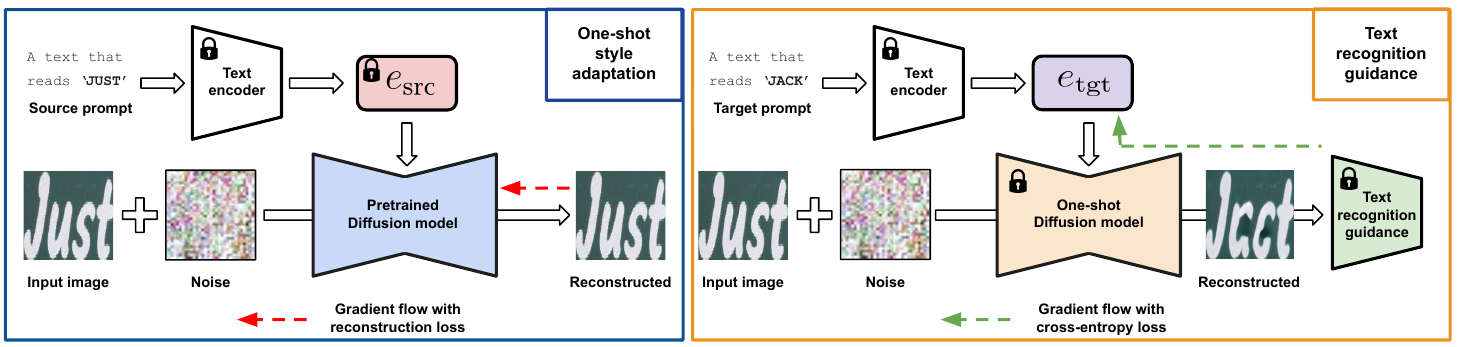}
    \end{center}
    \vspace{-0.5cm}
    \caption{
        The pipeline of our proposed method. The process is divided into 2 steps. One-shot style adaptation for fine-tuning the diffusion model and text recognition guidance for optimizing the target embedding. 
   }
\label{fig:framework}
\end{figure*}
\section{Proposed method}
Our proposed method is implemented based on the conditional LDM, which uses a pre-trained model from the large-scale text-to-image LAION dataset~\cite{SchuhmannArXiv2021}. However, generating scene texts remains challenging, as indicated in Fig.~\ref{fig:problem}. 
In this work, our training process comprises two stages. First, as a pre-requisite, we need to train (a.k.a fine-tune) the model on the text and edited-text pairs. To this end, we use a synthetic text generator (\ie, SynText~\cite{GuptaCVPR2016}) to produce scene text pairs. Through this strategy, the model can learn from ground-truth edited scenes in training. We only update the diffusion model $\epsilon_{\theta}$ on the SynText dataset while the condition model $\tau_{\psi}$ remains unchanged. Second, in order to manipulate scene text, we have two optimization processes, namely one-shot style adaptation and text recognition guidance as illustrated in Fig.~\ref{fig:framework}.


\subsection{One-shot style adaptation}
\label{sec:OSSA}
To manipulate a scene text, our model receives a source image $\Vec{x}$ and a text prompt $\Vec{y}$ which are mapped to the latent space $\Vec{z}$ and $\Vec{e}$, respectively. Our goal is to modify the source word to a target word while preserving the style of the source image. To this end, the model has to encode the characteristics of fonts, backgrounds, and colors. In the perspective of the text-to-image generation approach, this requires additional prompts from humans that might be difficult to find exact prompts. A sensible approach to address this problem is to invert the text from a source image~\cite{GalArXiv2022, RonArXiv2022}. 
However, we have observed that~\cite{KawarArXiv2022, RonArXiv2022} approaches
are suboptimal to maintain the original text style. For this reason, we propose a one-shot style adaptation strategy that directly optimizes the model's weights $\theta$ to achieve detail-preserved results :
    \begin{equation}
    \min_{\theta}\; \big\|\epsilon - \epsilon_{\theta}(\Vec{z}_t, t, \Vec{e}_{\mathrm{src}}) \big\|_2^2.
    \label{eq:oneshot}
    \end{equation} 
    Where $\Vec{z}_t$ is the source scene text in latent space.
    For each source image, we reset the parameters to initial $\epsilon_\theta$ and optimize the model using Eq.~\ref{eq:oneshot} yielding $\tilde{\epsilon}_{\theta}$. In each iteration, we randomly sample the noise $\epsilon$ in Eq.~\ref{eq:oneshot}. 
    
    There are significant differences with prior works for style adaptation or personalization. 
    Our style adaptation mechanism differs from DreamBooth~\cite{RuizArXiv2022} as it does not involve adding a unique identifier to the input text or generating extra images for class-specific prior preservation loss. Another strategy for personalization, known as textual inversion~\cite{GalArXiv2022}, may offer more efficient adaptation but often results in a loss of details. Furthermore, both strategies typically require more samples to learn the style. In contrast, our one-shot style adaptation only requires a single image for style adaptation and seamlessly incorporates input details.
    Imagic~\cite{KawarArXiv2022} requires optimizing the text embedding before fine-tuning the diffusion model, meaning that each edited prompt necessitates both embedding optimization and fine-tuning. In contrast, our method only requires training the diffusion model once. Even with an edited prompt, we only need to optimize the embedding without retraining the diffusion model.
    

\subsection{Text recognition guidance}
\begin{algorithm}
    \caption{Text Recognition Guidance}
    \begin{algorithmic}[1]
        \State {\textbf{Input :} Input latent image $\Vec{z}_0$, target embedding $\Vec{e}_{\mathrm{tgt}}$ , and target text $\mathcal{T}$}
        \State {\textbf{Output :} Optimized embedding $\tilde{\Vec{e}}_{\mathrm{tgt}}$} 
        \For{number of optimization iterations}
            \State {Get $t$ from Uniform($1, \cdots, T$)}
            \State {Generate $\epsilon$  from $\mathcal{N}(0, \mathbf{I})$}
            
            \State {Generate $\Vec{z}_{t} = \sqrt{\alpha_{t}}\Vec{z}_0 + \sqrt{1 - \alpha_{t}}\epsilon$)}
            
            \State {Obtain $\hat{\epsilon} = \tilde\epsilon_{\theta}(\Vec{z}_t, t, \Vec{e}_{\mathrm{tgt}})$}
            
            \State {Compute $\hat{\Vec{x}}_{edit}$ from Eq.~\ref{eq:rec_img}} 

            \State{Update ${\Vec{e}_{\mathrm{tgt}}}$ from Eq.~\ref{eq:nll_text}
            }
           
            
           
        \EndFor
   \end{algorithmic}
   \label{alg:tcl}
\end{algorithm}
\begin{figure*}[t]
    \centering
        \includegraphics[width=0.95\linewidth]{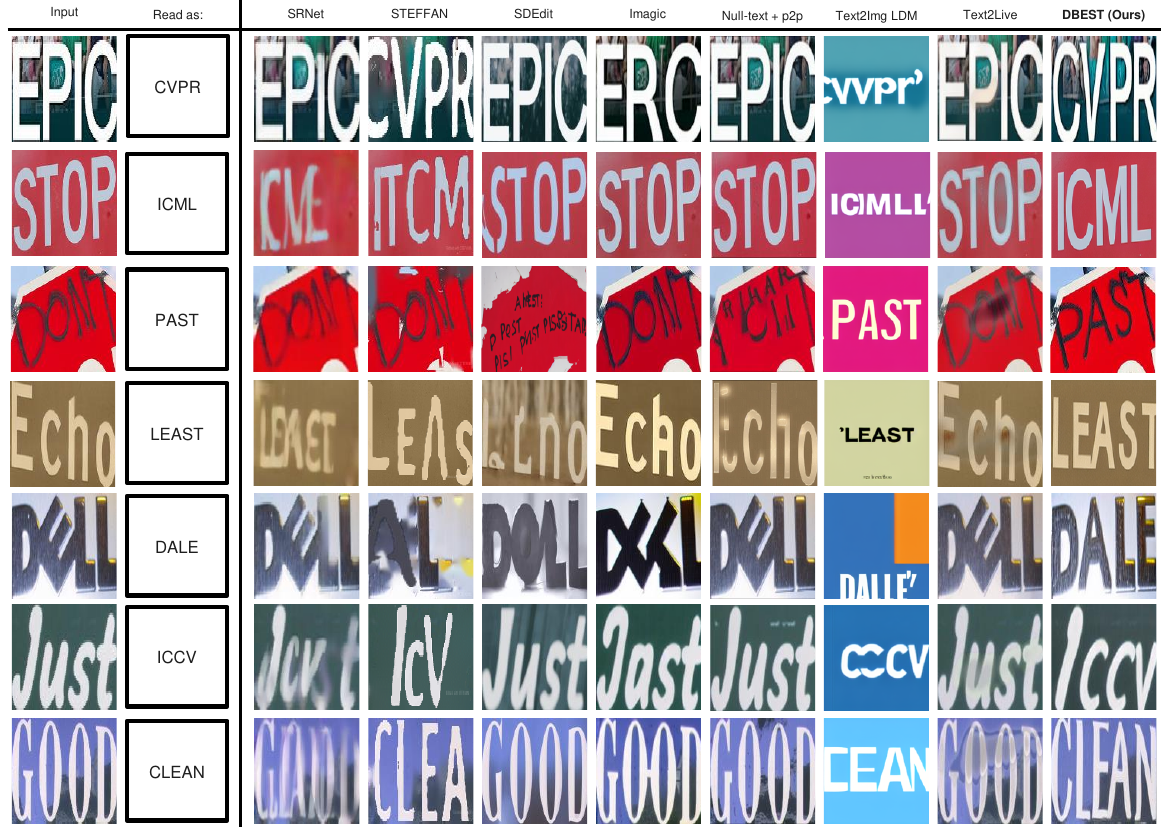}
    \centering
    \vspace{-0.2cm}
    \caption{Qualitative comparison on COCO-Text~\cite{VeitArXiv2016} and ICDAR2013~\cite{KaratzasICDAR2013} datasets.  DBEST (ours) achieves superior qualitative results compared to SRNet~\cite{WuACMM2019}, STEFFAN~\cite{RoyCVPR2020}, SDEdit~\cite{MengICLR2022}, Imagic~\cite{KawarArXiv2022}, Null-Inv~\cite{RonArXiv2022}+p2p~\cite{HertzArXiv2022}, Text2Img LDM~\cite{RombachCVPR2022}, Text2Live~\cite{BartalECCV2022}.  
    }
    \label{fig:itw}
\end{figure*}
    In addition to the style adaptation strategy, we aim to preserve the readability and performance on OCR. To this end, we introduce text recognition guidance integrated with the diffusion model. We use the style-adapted diffusion model $\tilde{\epsilon}_{\theta}$ as a starting point. In this stage, we start from the embedding of the target text $\Vec{e}_{\textrm{tgt}}$ and the text recognizer to guide the model to generate precise texts as shown in Fig.~\ref{fig:framework} (right). Firstly, as the diffusion process is performed in the latent space, 
    we need to reconstruct to the image level as followed:
    \begin{equation}
        \hat{\Vec{x}}_{edit} = \mathcal{D}(\hat{\Vec{z}}_0),
    \label{eq:rec_img}
    \end{equation} 
    where $\hat{\Vec{z}}_0$ is obtained by a denoising process for each $t \in \{1, \cdots, T\}$ as followed:
    \begin{equation}
        \hat{\Vec{z}}_0 =\frac{\Vec{z}_t - \sqrt{1 - \alpha_{t}}~{\tilde\epsilon_{\theta}}(\Vec{z}_t, t, \Vec{e}_{\mathrm{tgt}})}{\sqrt{\alpha_{t}}}.
    \label{eq:rec_latent}
    \end{equation} 
    The reconstructed scene text $\hat{\Vec{x}}_{edit}$ is fed into the text recognition model $f_\phi$ with the cross entropy loss $\mathcal{L}_{\mathrm{CE}}$. At this step, we keep the diffusion model unchanged as it has been adapted to the source style. Thus, we arrive at the textual inversion by employing iterative optimization on the target embedding $\Vec{e}_{\mathrm{tgt}}$ as: 
    \begin{equation}
        \tilde{\Vec{e}}_{\mathrm{tgt}} = \Vec{e}_{\mathrm{tgt}} - \nabla_{\Vec{e}_{\mathrm{tgt}}} \mathcal{L}_{\mathrm{CE}} (f_\phi, \hat{\Vec{x}}_{edit}, \mathcal{T}_{\mathrm{tgt}}), 
    \label{eq:nll_text}
    \end{equation} 
    where $\mathcal{T}_{\mathrm{tgt}}$ is the target text. 
    In our implementation, we set the minimum value of $t \geq 500$ to ensure the input image is mainly covered by $\epsilon$. The step-by-step of text recognition guidance is presented in Algorithm~\ref{alg:tcl}. 
\section{Experiments}
This section presents a comparison of our method with existing approaches for scene text manipulation, including those with and without a diffusion model. Also, we include ablation studies to understand each component of our proposed approach. To begin, we introduce the datasets, SOTA methods, and provide experimental details.
\vspace{-0.2cm}
\paragraph{Datasets.} We employ two types of datasets for our evaluation: synthesized and in-the-wild datasets. We generate the synthesized text scene dataset. The background and the text variation are obtained from SynText~\cite{GuptaCVPR2016}. Then, we follow the step-by-step procedures mentioned in SRNet and produce 100k training images. For evaluation data, the backgrounds are gathered from publicly available images from the internet, and various fonts are used to produce 600 images. The in-the-wild datasets we use are COCO-Text~\cite{VeitArXiv2016} and ICDAR2013~\cite{KaratzasICDAR2013}, which are frequently used for quantitative comparison. We randomly select 100 images for each dataset and choose the corresponding target text. For additional qualitative results, we apply our method to YouTube videos and the HierText dataset~\cite{LongCVPR2022} presented in the supplementary material.

\vspace{-0.4cm}
\paragraph{Baselines.} 

We compared our proposed method with both non-diffusion model methods, including Pix2pix~\cite{IsolaCVPR2017}, SRNet~\cite{WuACMM2019}, STEFANN~\cite{RoyCVPR2020}, Text2Live~\cite{BartalECCV2022}, and De-Render~\cite{ShimodaICCV2021}, and diffusion model-based methods, including Imagic~\cite{KawarArXiv2022}, text-to-image (Text2Img) LDM\cite{RombachCVPR2022}, Null-text~\cite{RonArXiv2022} + prompt-to-prompt (p2p)~\cite{HertzArXiv2022}, and SDEdit~\cite{MengICLR2022}. 

\begin{figure*}[t]
    \centering
        \includegraphics[width=0.95\linewidth]{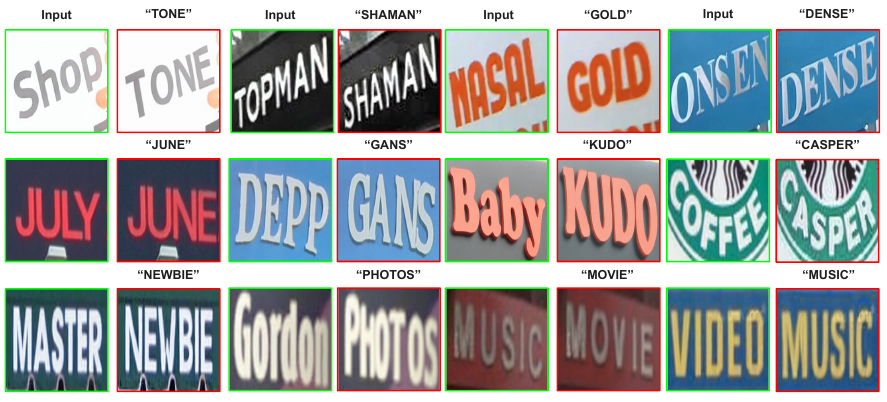}
   \caption{Result of text scene manipulation on row 1: ICDAR2015~\cite{KaratzasICDAR2015}, row 2: IIIT5K~\cite{MishraCVPR2012}, and row 3: SVT~\cite{PhanICCV2013}. The input image is represented by \textit{{\color{green} green box}} and the edited version by \textit{\color{red} red box}.}
\label{fig:itw_2}
\vspace{-0.5cm}
\end{figure*}

\vspace{-0.4cm}
\paragraph{Evaluation Measurements.} 
We evaluate the performance of the baselines using various measurements. Initially, we use the standard evaluation metrics for image manipulation, including PSNR, SSIM, and LPIPS~\cite{ZhangCVPR2018} scores on the SynText dataset. We compare the models based on character recognition benchmarks. For this purpose, we utilize a pre-trained weight of scene text recognition~\cite{YanCVPR2021} to determine if the generated sample is readable. We measure the recognition performance at both the character level and word level. 

\subsection{Evaluations}
\begin{table}[t]
    \centering
       \vspace{0.2cm}
    \setlength{\extrarowheight}{5pt}
     \resizebox{0.49\textwidth}{!}{
    \Large\addtolength{\tabcolsep}{0.1pt}
            \begin{tabular}{l  c  c  c  c  c}
                \hline
                \multirow{2}{*}{Methods} & \multirow{2}{*}{PSNR ($\uparrow$)} & \multirow{2}{*}{SSIM ($\uparrow$)} & \multirow{2}{*}{LPIPS ($\downarrow$)} & \multicolumn{2}{c}{OCR Acc.~$(\% \uparrow)$}\\
                \cline{5-6}
                & & & & char & word\\
                \hline
                    Pix2pix~\cite{IsolaCVPR2017}& 28.39 & 0.47 & 0.52 & 40.83 & 9.50 \\
                    SRNet~\cite{WuACMM2019} & 29.21 & 0.49 & 0.53 & 88.70 & 71.16\\
                    De-Render*~\cite{ShimodaICCV2021} & 29.68 & \textbf{0.65} & 0.48 & 40.82 & 17.39\\
                    SDEdit~\cite{MengICLR2022} & 29.35 & 0.45 & 0.44 & 4.91 & -\\
                    LDM~\cite{RombachCVPR2022} & 29.78 & 0.51 & 0.38 & 44.33 & 11.50\\
                    Imagic~\cite{KawarArXiv2022} & 28.18 & 0.30 & 0.61 & 77.75 & 50.66\\
                    Imagic with SynText~\cite{KawarArXiv2022} & 28.33 & 0.31 & 0.48 & 94.33 & 82.33\\
                    
                \rowcolor{blond}
                    DBEST~(Ours) & \textbf{30.09} & 0.54 & \textbf{0.29} & \textbf{94.58} & \textbf{84.83}\\
                \hline
            \end{tabular}
        }
    \caption{The average of PSNR, SSIM, LPIPS, and OCR scores on SynText dataset. The notations '-' and '*' denote a method with no correct word and unfair comparison, respectively. The best score is denoted by \textbf{bold} text. }
    \label{table:syntext_eval}
    \vspace{-0.3cm}
\end{table}
\begin{table}[t]
    \centering
       \vspace{0.2cm}
    \setlength{\extrarowheight}{2pt}
     \resizebox{0.47\textwidth}{!}{
    \Large\addtolength{\tabcolsep}{5pt}
            \begin{tabular}{l  c  c  c  c }
                \hline
                
                \multirow{3}{*}{Methods} & \multicolumn{4}{c }{OCR Acc.~$(\% \uparrow)$}\\
                & \multicolumn{2}{c }{COCO-Text} & \multicolumn{2}{c }{ICDAR2013} \\
                \cline{2-5}
                & char & word & char & word\\
                \cline{1-5}
                    Pix2pix~\cite{IsolaCVPR2017} & 20.00 & - & 17.60 & - \\
                    SRNet~\cite{WuACMM2019} & 56.58 & 21.00 & 51.79 & 25.51 \\
                    De-Render*~\cite{ShimodaICCV2021} & 24.63 & 2.00 & 18.54 & - \\
                    Imagic~\cite{KawarArXiv2022} & 20.00 & 1.00 & 14.70 & - \\
                    SDEdit~\cite{MengICLR2022} & 17.31 & - & 18.30 & - \\
                    Null-Inv~\cite{RonArXiv2022} + P2P~\cite{HertzArXiv2022} & 18.78 & - & 17.37 & - \\
                    \rowcolor{blond}
                    DBEST~(Ours) & \textbf{94.15} & \textbf{83.00} & \textbf{98.12} & \textbf{92.00} \\
                    
                \hline
            \end{tabular}
        }
    \caption{The average OCR accuracy on COCO-Text and ICDAR2013 dataset. The notations '-' and '*' denote a method with no correct word and unfair comparison, respectively. The best score is denoted by \textbf{bold} text.}
    \label{table:itw}
    \vspace{-0.4cm}
\end{table}
\begin{figure*}[t]
    \centering
        \includegraphics[width=0.95\linewidth]{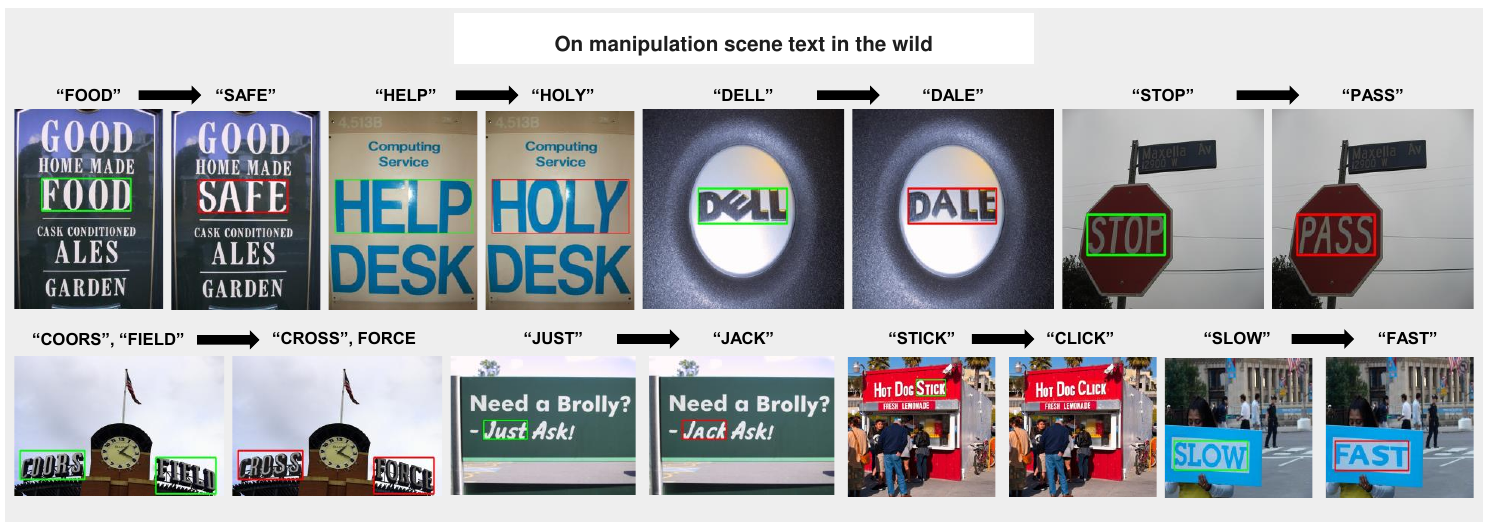}
   \caption{Given a single word \textit{{\color{green} green box}} from in-the-wild images and the desired text, our method successfully to edit the text with the desired text in the image \textit{{\color{red} red box}}.}
   \label{fig:teaser}
   \vspace{-0.4cm}
\end{figure*}

\paragraph{State-of-the-art comparison.} 
    We conduct comparisons among state-of-the-art (SOTA) methods and found that SRNet is the best-performing method among them. In Table~\ref{table:syntext_eval}, we compare our method with Pix2pix, SRNet, De-Render, SDEdit, Imagic, and LDM as the SOTA methods on SynText dataset. In Table~\ref{table:itw}, we compare our method with SRNet, De-Render, Imagic, SDEdit, and Null-text + P2P as the SOTA methods on in-the-wild datasets.   
    Our approach achieved SOTA performance on image quality assessments. Furthermore, in Table~\ref{table:syntext_eval}, we achieve 94\% and 84\% for character and word recognition accuracy, respectively. Even when evaluating in-the-wild data (as shown in Table~\ref{table:itw}), our method maintained consistent performance, whereas SRNet's performance significantly decreased as well as Imagic's performance. In Fig.~\ref{fig:itw}, it is demonstrated that SDEdit, Imagic, Null-text, Text2Img LDM, and Text2live are unable to modify the source image. The main reason is that language models, such as BERT or CLIP, do not possess adequate knowledge of scene text. 
    Meanwhile, STEFFAN heavily relies on per-character map estimation, and SRNet struggles with reconstruction due to the diverse characteristics of in-the-wild datasets compared to SynText. Our method, on the other hand, can manipulate the scene while preserving the source style. However, when we applied De-Render to both evaluation tables, we observed that several images were undetected. This is because this method is limited to digital documents. 
\begin{table}[t]
    \centering
       \vspace{0.2cm}
    \setlength{\extrarowheight}{2pt}
     \resizebox{0.49\textwidth}{!}{
    \Large\addtolength{\tabcolsep}{0.5pt}
            \begin{tabular}{c  c  c  c  c  c  c}
                \hline
                \multirow{2}{*}{SynText} & \multirow{2}{*}{Text} & \multirow{2}{*}{PSNR ($\uparrow$)} & \multirow{2}{*}{SSIM ($\uparrow$)} & \multirow{2}{*}{LPIPS ($\downarrow$)} & \multicolumn{2}{c}{OCR Acc.~$(\% \uparrow)$}\\
                \cline{6-7}
                & & & & & char & word\\
                \hline
                    \color{red}\xmark & \color{red}\xmark & 29.78 & 0.51 & 0.38 & 44.33 & 11.50\\
                    \color{red}\xmark &  \color{green}\cmark & 29.90 & 0.52 & 0.35 & 76.70 & 50.00\\
                    
                    \color{green}\cmark & \color{red}\xmark & \textbf{30.09} & \textbf{0.54} & 0.30 & 84.25 & 60.16\\

                    \color{green}\cmark & \color{green}\cmark & \textbf{30.09} & \textbf{0.54} & \textbf{0.29} & \textbf{94.58} & \textbf{84.83}\\
                \hline
            \end{tabular}
        }
   
    \caption{Ablation studies on the SynText dataset. The best score is denoted by \textbf{bold} text. The column 'SynText' and 'Text' denote a utilization of SynText dataset and text recognition guidance, respectively.}
    \label{table:ablations}
    \vspace{-0.6cm}
\end{table}

\vspace{-0.3cm}
\paragraph{Result on the in-the-wild dataset.} 
    Fig.~\ref{fig:itw_2} illustrates more examples of scene text manipulation on ICDAR2015~\cite{KaratzasICDAR2015}, IIIT5K~\cite{MishraCVPR2012}, and SVT~\cite{PhanICCV2013} datasets.  ICDAR2015 and SVT datasets are particularly challenging due to the presence of factors that can affect captured scene text, including geometry distortion, noise effects, and motion blur. Even with those conditions, our method is still able to manipulate the scene text. 
    In addition, our method's ability to manipulate text in a full image is demonstrated in Fig.~\ref{fig:teaser}. To accomplish this, we can utilize a scene text detection algorithm like EAST~\cite{ZhouCVPR2017} or a provided text bounding box to identify each text instance within the image. Then, each identified text area is extracted and provided to our method along with the target text. Finally, the edited image is projected back onto the original image to generate the desired output.
\begin{figure}[t]
    \begin{center}
        \includegraphics[width=1.\linewidth]{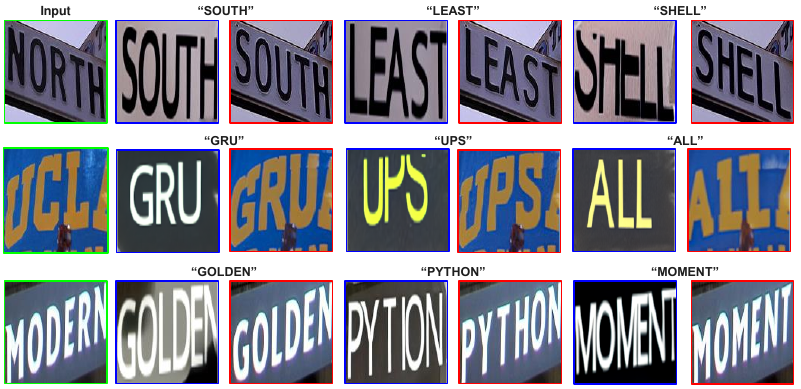}
    \end{center}
    \vspace{-0.4cm}
    \caption{Result from a single image with different target texts. 
    The input image is represented by \textit{{\color{green} green box}}, the edited image without one-shot diffusion model is shown by \textit{{\color{blue} blue box}} and the final edited version by \textit{\color{red} red box}.} 
    \label{fig:one_2_many}
    \vspace{-0.4cm}
\end{figure}

\subsection{Ablation studies}
Below, we investigate how our method behaves and the impacts of each component in our work.
\vspace{-0.2cm}
\paragraph{Why do we need pre-trained with SynText dataset?}
At its core, our problem is the inability of text-to-image models to handle the scene text domain, as depicted in the bottom part of Fig.~\ref{fig:problem}. By and large, we observe that the LAION dataset~\cite{SchuhmannArXiv2021} comprises general scenes. Therefore, a straightforward approach of using LDM for text-to-scene text fails. In Table~\ref{table:ablations}, we present significant improvement in terms of OCR accuracy, where the first row represents the accuracy of text-to-image LDM as a baseline.
\begin{figure}[t]
    \begin{center}
        \includegraphics[width=1.\linewidth]{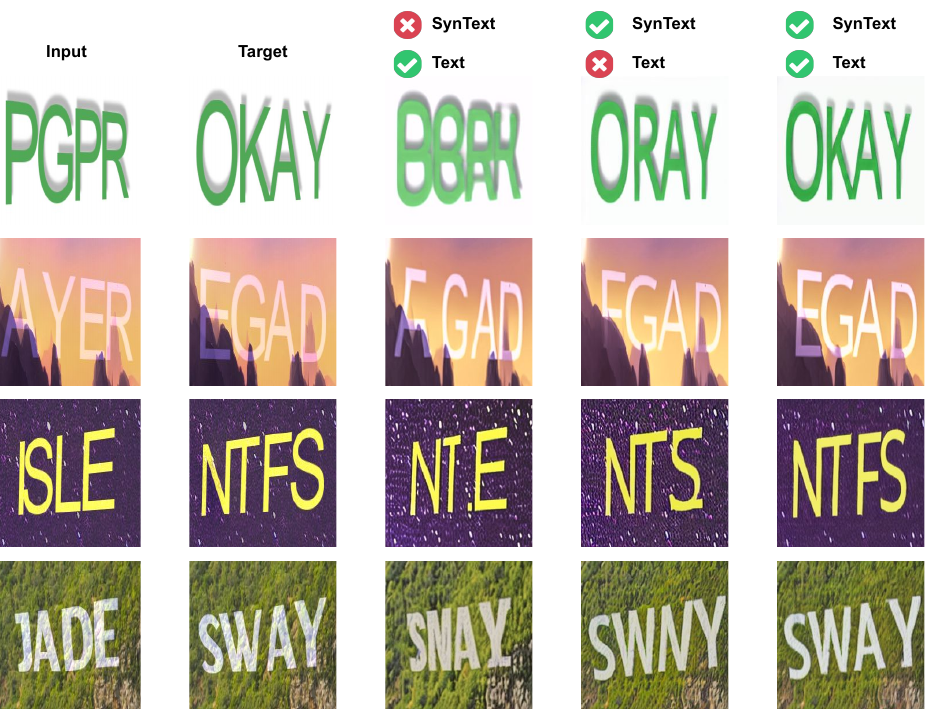}
    \end{center}
    \vspace{-0.5cm}
    \caption{Ablation result of our method on Syntext dataset. 
    The label 'SynText' and 'Text' denote a utilization of SynText dataset and text recognition guidance, respectively.
    }
    \label{fig:ablation}
    \vspace{-0.5cm}
\end{figure}

\begin{figure}[t]  
    \begin{center}
        \includegraphics[width=1.\linewidth]{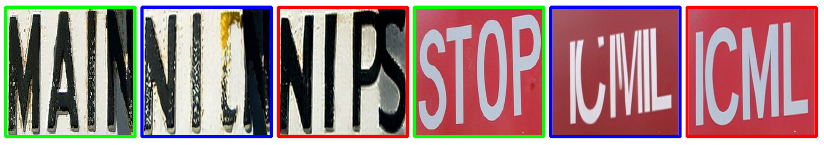}
    \end{center}
    \vspace{-0.5cm}    
        \caption{`MAIN', `STOP' $\xrightarrow{}$ `NIPS', `ICML'. The input image is represented by \textit{{\color{green} green box}}, the result by 'Imagic with SynText' is shown by \textit{{\color{blue} blue box}} and ours by \textit{\color{red} red box}.} 
\label{fig:imagicsyntext}
\vspace{-0.3cm}
\end{figure}

\paragraph{Why do we need one-shot style adaptation?}
The one-shot style adaptation for the diffusion model is crucial for maintaining the source appearance and text geometry of the edited text. As shown in Fig.~\ref{fig:one_2_many}, our approach can successfully edit the text, but the edited result loses its source appearance without the one-shot method. In other words, without the one-shot diffusion model, our method may generate an image that does not resemble the original input. The one-shot style adaptation reduces the distribution space of the diffusion model, ensuring that the generated image maintains the source appearance of the input image.
    Furthermore, our method offers the advantage of allowing multiple text editing tasks to be performed with just a single one-shot style adaptation. This means that users can easily modify various pieces of text while maintaining the same style without requiring additional style adaptation. The effectiveness of this can be observed in Fig.~\ref{fig:one_2_many}.
\vspace{-0.3cm}
\paragraph{Why do we need text recognition guidance?}
According to Table.~\ref{table:ablations}, when text recognition guidance is not used, the character accuracy and word accuracy are 84.25 \% and 60.16 \%, respectively. This is further confirmed by Fig.~\ref{fig:ablation}, where we can see that the manipulated scene text contains some character errors. However, with the addition of text recognition guidance, the diffusion model is guided to revise the character errors, resulting in a significant improvement in word accuracy, as shown in Table.~\ref{table:ablations}.
In essence, the text recognition guidance helps to improve the accuracy of the edited text scene by guiding the diffusion model to revise the character errors. By incorporating this guidance, our approach is able to generate more readable and accurate edited text scenes, improving the overall quality of the output.
\vspace{-0.3cm}
\paragraph{What is the key difference with Imagic~\cite{KawarArXiv2022} ?}
We emphasize that Imagic requires interpolating between the source and target embeddings with an associated parameter difficult to be finetuned. While, a simple Imagic model trained on Syntext denoted as 'Imagic with SynText' failed to edit scene texts effectively (e.g., inconsistent styles) as depicted in Fig.~\ref{fig:imagicsyntext} (col. 2) and low score on image quality performance in Table~\ref{table:syntext_eval}. In contrast, our proposed method can easily generate well-accepted results by only optimizing the target text embedding without an extra interpolation parameter. 

\begin{figure}[t]
    \centering
        \includegraphics[width=1.02\linewidth]{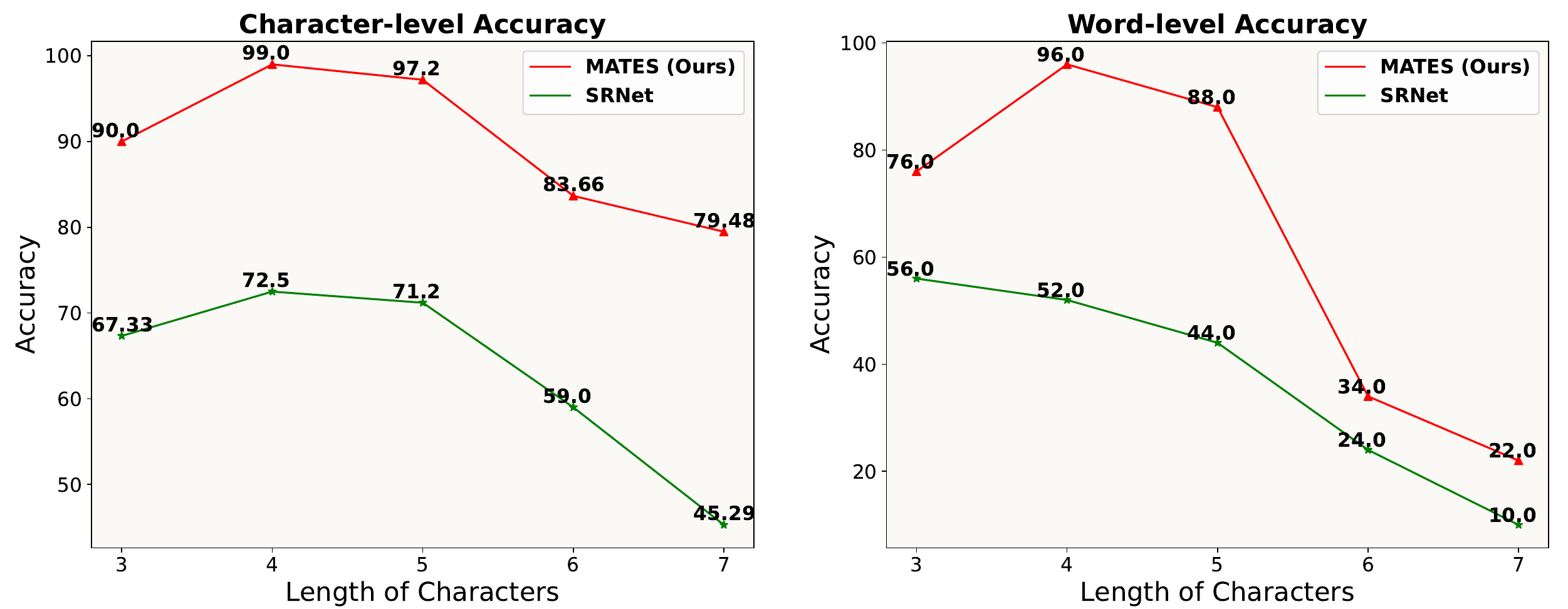}
    \caption{Analyzing the edited result with various length of characters on the in-the-wild dataset. The input image is constrained to 4 length of characters and the target text is varying from 3-7 characters. Though the target text has more characters, our method is still superior compared to SRNet in terms of accuracy in character and word levels. 
    } 
    \label{fig:loc_plot}
    \vspace{-0.5cm}
\end{figure}

\vspace{-0.3cm}
\paragraph{Length of characters.} 
We analyze the effect of character length on text manipulation performance by using an input image with four different character lengths. As shown in Fig.~\ref{fig:loc_plot}, the results indicate a significant decrease in performance when manipulating text with more than five characters. We attribute this drop in performance to the model's inability to adjust the font size based on the given text and the available space in the scene.
Currently, our model lacks the ability to adapt the font size based on the text length, which limits its performance in manipulating longer texts. This limitation is evident in Fig.~\ref{fig:loc_plot}, where the model struggles to maintain accuracy when manipulating text with more than five characters. Therefore, improving the model's ability to adjust font size dynamically based on the text length and available space in the scene could enhance its performance in manipulating longer texts.

\section{Conclusion}
    We propose a diffusion model for scene text manipulation with one-shot style adaptation and text recognition guidance. In our evaluation, we use synthesized and in-the-wild datasets, which have different characteristics. As our experimental evaluations indicate, our methods comfortably outperform the existing methods in both datasets. Furthermore, the ablation studies we conducted indicate that each process within our method has a distinct task and is crucial to its overall performance.


Our method currently has two limitations. First, the inference time is around 8 minutes per image, making it unsuitable for real-time applications. Second, there is a constraint on the character length that can be manipulated. As mentioned, adjusting the font size is another challenge we aim to address in future work. 
\vspace{-0.3cm}
\paragraph{Acknowledgement.} We would like to express our gratitude to Samuel Timotius Adi for providing the resources necessary to complete this paper. 
{\small
\bibliographystyle{ieee_fullname}
\bibliography{egbib}
}
\twocolumn[{
    \begin{center}
        \Large
        \textbf{{** Supplementary material ** \\ On Manipulating Scene Text in the Wild with Diffusion Models}}
    \end{center}
}]

In this supplementary material,
we provide (i) more details about our implementation setting and the SynText dataset used in our experiments (e.g.,
dataset creation),  (ii) additional experiments on one-shot ablation studies, and (iii) additional qualitative results on ICDAR, COCO-Text, SVT, IIIT5K, and HierText datasets as well as Youtube video. 

\section{Implementation Setting}
We began by setting up the pre-trained LDM model, which uses text-to-image technology~\cite{RombachCVPR2022}, and incorporated the ABINet text recognition model~\cite{FangCVPR2021}. Our training process involved utilizing the Syntext dataset for 500k iterations, with a learning rate of $1e^{-6}$ and a batch size of 1. It's important to note that we only trained the diffusion model while keeping the language model unchanged from its original state. To manipulate the text in the input image, we fine-tuned the diffusion model using a given image for 1500 iterations, with a learning rate of $1e^{-6}$. Next, we optimized the target embedding using cross-entropy loss over 1000 iterations, with a learning rate of $1e^{-4}$. We employed an ADAM optimizer and performed  on a single RTX 3090.

\section{Synthesized Dataset}
Our main paper briefly explains that our synthesized scene texts are generated using SynText~\cite{GuptaCVPR2016}. Below, we provide a list of steps to build the synthesized text pairs, which include both the source and target texts:

\renewcommand{\labelenumi}{\roman{enumi}.}
\begin{enumerate}
    \item{
        We prepare source texts, target texts, and background images from  SynText~\cite{GuptaCVPR2016}. 
    }
    \item{
        We randomize the augmentation parameters such as font type \& size, text geometry, color augmentation, and text effects as shown in Table.~\ref{tabel:augm} as well as the desired image resolution.   
    }
    \item{
         Both texts are rendered from the text space to the image space using the Pygame~\cite{Pygame} library, and we apply various augmentations such as font type, font size, curve text rate, underline text rate, strong text rate, and oblique rate. During this step, we apply a standard color to the text, such as black.
    }
    \item{
        The rendered text is additionally augmented with geometry augmentation, such as zoom rate, shear rate, rotation rate, and perspective rate, to supplement the text location and perspective. 
    }
    \item{ 
        The background is randomly cropped according to the desired resolution, and then several augmentations are applied, including brightness, contrast, color, and additional padding adjustments. 
    }
    \item{
        The rendered text and pre-processed background are then integrated. During this process, we apply text effect augmentation \eg, shadow effects and text colors. Lastly, we combine text and background using the Poisson blending algorithm~\cite{PatrickACMTG2003}.
    }
\end{enumerate}
We slightly modified the augmentation parameters for the evaluation data by setting the minimum size to 128 for each width and height to ensure high-quality images. We also added more font types, such as DecaySans and Chickenic. Finally, we adjusted the contrast and brightness values to validate that the recognition model could recognize the text correctly.

\begin{table*}[t] 
    \centering
       \vspace{0.2cm}
    \setlength{\extrarowheight}{2pt}
     \resizebox{0.85\textwidth}{!}{
            \begin{tabular}{l  p{2.cm} p{2.0 cm}  c c c c c c c c }
                \hline
                \multirow{2}{*}{Types} & \multirow{2}{*}{Apply} & \multirow{2}{*}{Aug. Method} & \multicolumn{7}{ c }{Values}\\
                 \cline{4-10}
                 &  &  & Min & Max & Rate & Scale & Grid size & Mag. & Desc.\\
                \hline
                \multirow{5}{*}{Color} & \multirow{4}{*}{Background} & Color & 0.7 & 1.3 & 0.8 & - & - & - & -\\
                 &  & Brightness & 0.7 & 1.5 & 0.8 & - & - & - & -\\
                 &  & Contrast & 0.7 & 1.3 & 0.3 & - & - & - & -\\
                 &  & Resolution & 64 & - & - & - & - & - & -\\
                & Text & Color & 0.7 & 1.3 & 0.8 & - & - & - & -\\
                \hline      
                \multirow{6}{*}{Font} & \multirow{6}{*}{Text} & Size & 25 & 60 & - & - & - & - & -\\
                 &  & \multirow{2}{*}{Type} & \multirow{2}{*}{-} & \multirow{2}{*}{-} & \multirow{2}{*}{-} &\multirow{2}{*}{-} & \multirow{2}{*}{-} & \multirow{2}{*}{-} & Arial \& \\
                 &  &  &  &  &  &  &  &  & OpenSans \\
                &  & Underline & - & - & 0.01 & - & - & - & -\\
                &  & Strong & - & - & 0.07 & - & - & - & -\\
                &  & Oblique & - & - & 0.02 & - & - & - & -\\
                \hline
               \multirow{6}{*}{Geometry} & \multirow{6}{*}{Text} & Zoom & - & - & - & 0.1 & - & - & -\\
               &  & Rotate & - & - & - & 1 & - & - & -\\
               &  & Shear & - & - & - & 2 & - & - & -\\
               &  & Perspective & - & - & - & 0.0005 & - & - & -\\
               &  & Elastic & - & - & 0.001 & - & 4 & 2 & -\\
               &  & Curve & - & - & - & 0.05 & 0.1 & - & -\\
               \hline
               \multirow{2}{*}{Effects } & \multirow{2}{*}{Text} & Border & - & - & 0.02  & -& - & - & -\\
               &  & Shadow & - & - & 0.02 & - & - & - & -\\
               
                \hline
                \hline
            \end{tabular}
        }
    \caption{The list of augmentations for generating the SynText dataset. ('Types': type of augmentation, 'Apply': target augmentation, 'Aug. Method': specific augmentation method, 'Min': minimum value, 'Min': maximum value, 'Rate': the possibility the augmentation is applied, 'Scale': the scale value, 'Grid size': the grid size for elastic augmentation, 'Mag': the magnitude value, and 'Desc.': is the description.  }
     \vspace{-0.2cm}
    \label{tabel:augm}
\end{table*} 

\section{Additional Experiments}
In this section, we provide more analysis to ablate our framework using one-shot style adaptation. Moreover, we also show the trade-off analysis between style preservation and text editing. We present more qualitative results on various datasets \eg, ICDAR2013~\cite{KaratzasICDAR2013}, ICDAR2015~\cite{KaratzasICDAR2015}, COCO-Text~\cite{VeitArXiv2016}, SVT~\cite{PhanICCV2013}, IIIT5K~\cite{MishraCVPR2012}, HierText~\cite{LongCVPR2022}, and Youtube videos.  

\begin{table}[t] 
    \centering
       \vspace{0.2cm}
    \setlength{\extrarowheight}{2pt}
     \resizebox{0.49\textwidth}{!}{
    \Large\addtolength{\tabcolsep}{0.5pt}
            \begin{tabular}{c  c  c  c  c  c  c}
                \hline
                \multirow{2}{*}{One-shot} & \multirow{2}{*}{Text} & \multirow{2}{*}{PSNR ($\uparrow$)} & \multirow{2}{*}{SSIM ($\uparrow$)} & \multirow{2}{*}{LPIPS ($\downarrow$)} & \multicolumn{2}{c}{OCR Acc.~$(\% \uparrow)$}\\
                \cline{6-7}
                & & & & & char & word\\
                \hline
                    \color{red}\xmark & \color{red}\xmark & 28.62 & 0.39 & 0.54 & 92.50 & 77.50\\
                    \color{red}\xmark &  \color{green}\cmark & 28.67 & 0.37 & 0.55 & \textbf{96.79} & \textbf{90.50}\\
                    \color{green}\cmark & \color{red}\xmark & \textbf{30.09} & \textbf{0.54} & 0.30 & 84.25 & 60.16\\
                    \color{green}\cmark & \color{green}\cmark & \textbf{30.09} & \textbf{0.54} & \textbf{0.29} & 94.58 & 84.83\\
                \hline
            \end{tabular}
        }
   
    \caption{Ablation studies on the SynText dataset. The best score is denoted by \textbf{bold} text. The column 'One-shot' and 'Text' denote the utilization of one-shot style adaptation method and text recognition guidance, respectively.}
    \label{table:oneshot}
\end{table}

\begin{figure}[t]
    \centering
        \includegraphics[width=1.\linewidth]{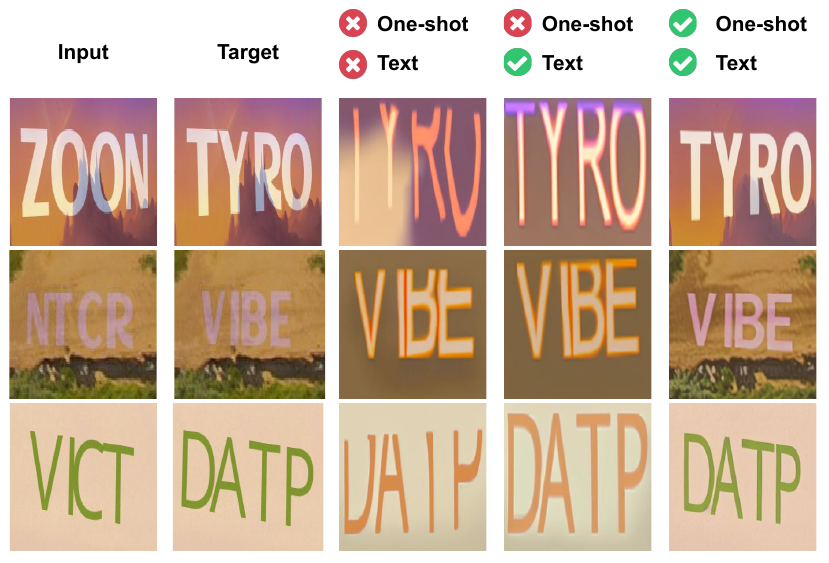}
    \caption{Ablation result of our method on Syntext dataset. 
    The label 'One-shot' and 'Text' denote the utilization of SynText dataset and text recognition guidance, respectively.
    } 
    \label{fig:onsa_abl}
    \vspace{-0.5cm}
\end{figure}

\paragraph{One-shot style adaptation.}
In Table.~\ref{table:oneshot}, we show the importance of one-shot style adaptation. We use the diffusion model $\epsilon_{\theta}$ fine-tuned on our created synthetic scene texts. Although the OCR score is slightly lower, the image quality score significantly drops. Without one-shot style adaptation, the scores plummet $1.47$, $0.15$, and $0.25$ for PSNR, SSIM, and LPIPS scores, respectively, compared to our complete framework. It is clearly shown in Fig.~\ref{fig:onsa_abl} where the style differs entirely from the source image. Even though we apply text recognition guidance, it only revises the text content, not the overall style.  

\paragraph{Preserving style trade-off.} We observe the trade-off between preserving style and editability in Table \ref{table:oneshot}. The first row represents the results without a one-shot step. The OCR score is high, but the image assessment score is significantly lower. In contrast, the third row shows a considerable increase in image assessment when we apply the one-shot approach, but at the cost of a considerably lower OCR score.

\begin{figure*}[t]
    \centering
        \includegraphics[width=1.\linewidth]{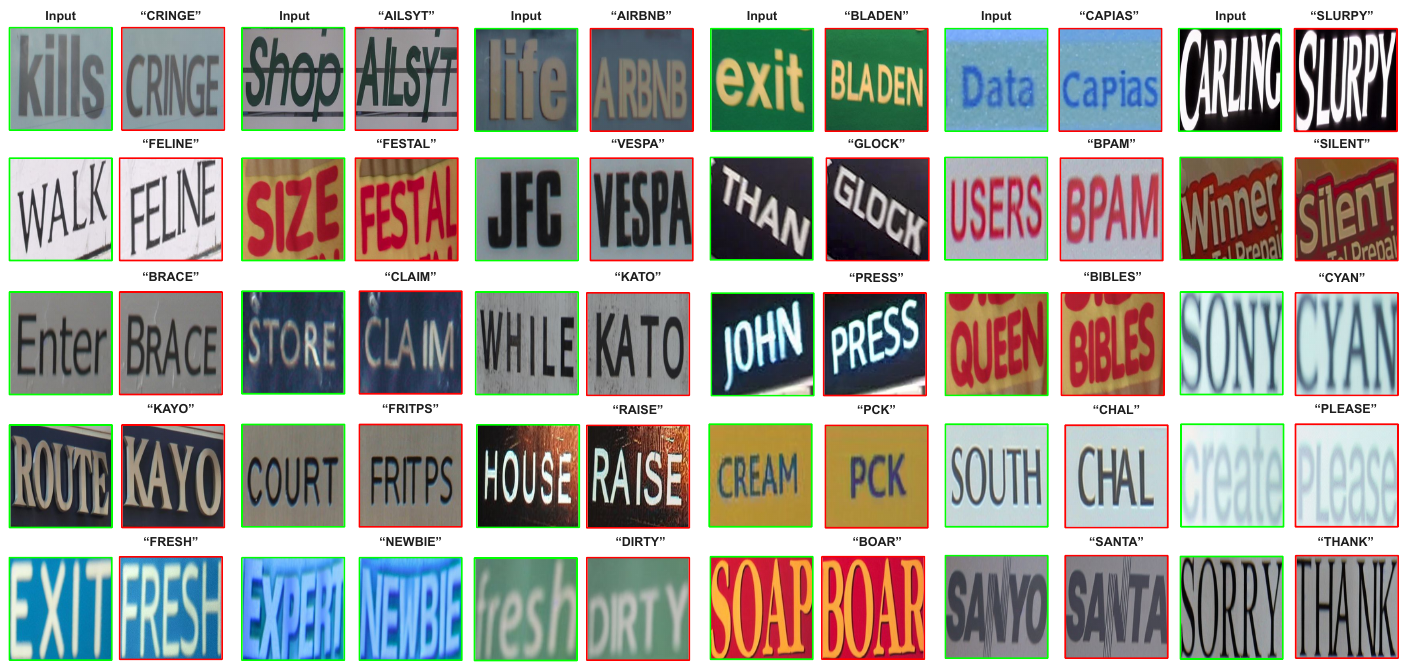}
   \caption{
       Given a single word \textit{{\color{green} green box}} from in-the-wild images and the desired text, our method successfully edits the text with the desired text in the image \textit{{\color{red} red box}} in mixed datasets such as ICDAR2013~\cite{KaratzasICDAR2013}, ICDAR2015~\cite{KaratzasICDAR2015}, COCO-Text~\cite{VeitArXiv2016}, and SVT~\cite{PhanICCV2013}.
   }
   \label{fig:itw_add}
   \vspace{-0.1cm}
\end{figure*}

\begin{figure*}[t]
    \centering
        \includegraphics[width=1.\linewidth]{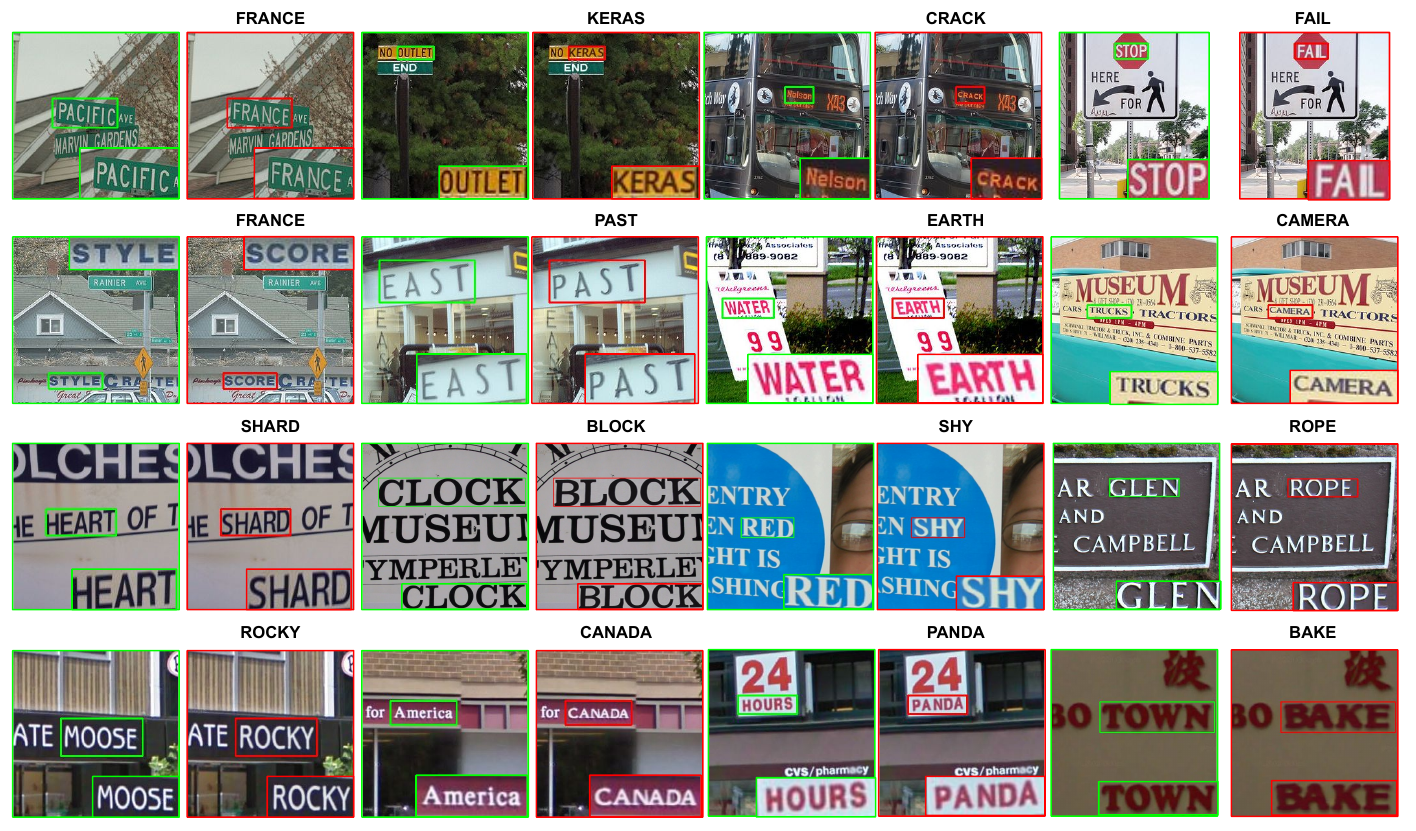}
   \caption{
       Given a cropped single word from in-the-wild images using a bounding box marked by a \textit{{\color{green} green box}}, and specifying the desired text, our method successfully edits the text to match the desired text and can replace the original word in the original image, as shown by \textit{{\color{red} red box}}.
   }
   \vspace{-0.2cm}
   \label{fig:itw_real}
\end{figure*}

\begin{figure*}[t]
    \centering
        \includegraphics[width=1.\linewidth]{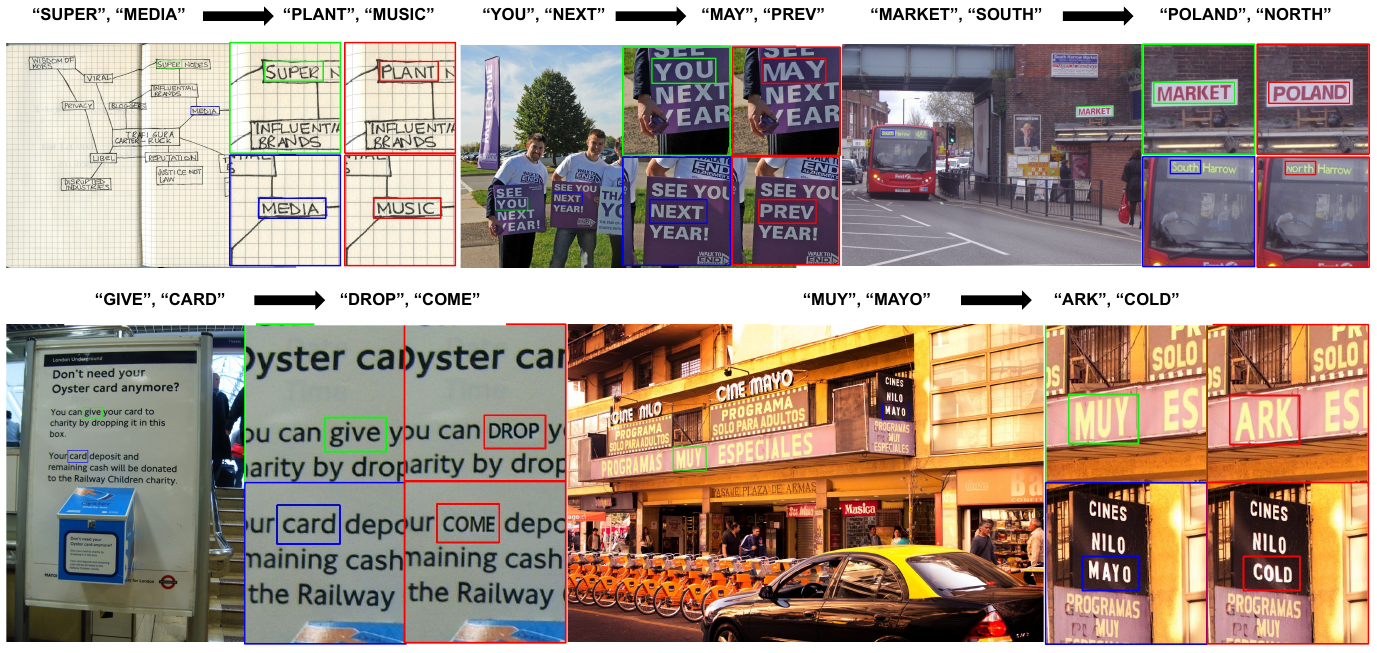}
   \caption{
       Given a cropped single word from HierTextt~\cite{LongCVPR2022} images using a bounding box marked by \textit{{\color{green} green box}} and \textit{{\color{blue} blue box}}, and specifying the desired text, our method successfully edits the text to match the desired text and can replace the original word in the original image, as shown by \textit{{\color{red} red box}}.
   }
   \vspace{-0.2cm}
   \label{fig:hiertext}
\end{figure*}

\begin{figure*}[t]
    \centering
        \includegraphics[width=1.\linewidth]{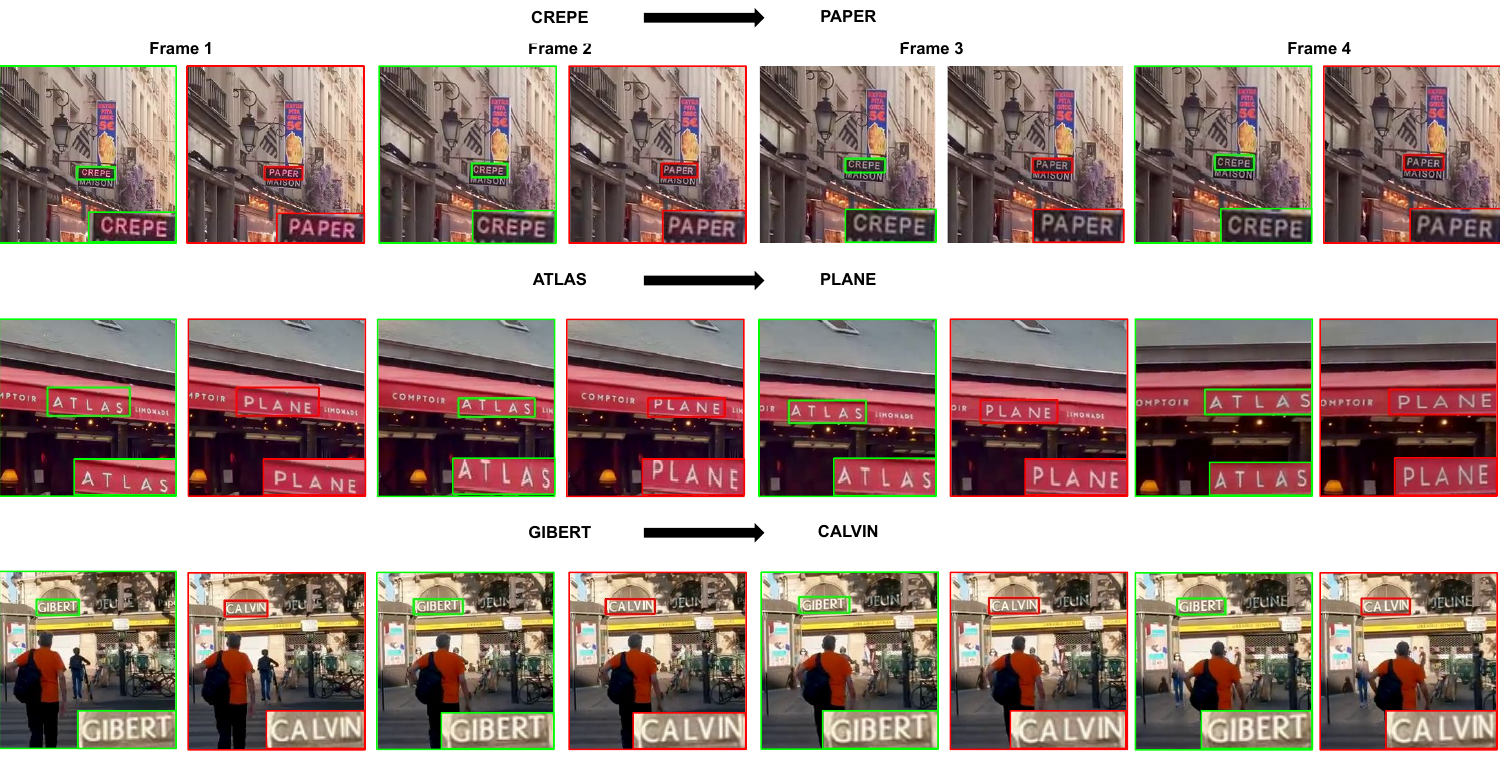}
   \caption{
       Given a cropped single word from Youtube frames using a bounding box marked by a \textit{{\color{green} green box}}, and specifying the desired text, our method successfully edits the text to match the desired text and can replace the original word in the original image, as shown by \textit{{\color{red} red box}}.
   }
   \vspace{-0.2cm}
   \label{fig:youtube}
\end{figure*}

\paragraph{Qualitative results}
We present additional qualitative results for our method in Fig.~\ref{fig:itw_add}, Fig.~\ref{fig:itw_real} on benchmark datasets, including ICDAR2013~\cite{KaratzasICDAR2013}, ICDAR2015~\cite{KaratzasICDAR2015}, COCO-Text~\cite{VeitArXiv2016}, SVT~\cite{PhanICCV2013}, and IIIT5K~\cite{MishraCVPR2012}. In Fig.~\ref{fig:itw_add}, we add more examples where the target text is more varied in terms of length of characters. Furthermore, in Fig.~\ref{fig:itw_real}, we show that our method is able to replace the source scene text in the original image with the edited version while preserving the visual characteristics of the image, such as its geometry and style. We also demonstrate our result on the recent scene text dataset namely HierText~\cite{LongCVPR2022} in Fig.~\ref{fig:hiertext} and more real-case such as Youtube videos in Fig.~\ref{fig:youtube}. For the YouTube video collection, we focused on travel-related content. We extracted frames from these videos using FFmpeg and selected frames that contained scene text. Note that these collected frames posed additional challenges:(1) The quality of the frames is uncontrollable (\eg it can be noisy and/or blurry due to the recording device and motion blur). (2) Since we specifically chose travel videos, the size of the scene text is typically quite small. Despite these challenges, our method successfully replaces the source scene text while preserving the characteristics, as demonstrated in Fig.~\ref{fig:youtube}.

\section{Societal Impact}
Our project aims to solve scene text manipulation for text translation and obscure sensitive information. However, we recognize that our work may also have unintended consequences, such as creating and disseminating false information. To address this issue, we are developing a fake scene text detection system that can help identify and prevent the creation of fake documents that use manipulated text. 

\end{document}